\documentclass[letterpaper]{article} 
\usepackage{aaai24}  
\usepackage{times}  
\usepackage{helvet}  
\usepackage{courier}  
\usepackage[hyphens]{url}  
\usepackage{graphicx} 
\usepackage{natbib}  
\usepackage{caption} 
\usepackage{algorithm}
\usepackage{algorithmic}
\usepackage{xspace}
\usepackage{makecell}
\usepackage{booktabs}
\usepackage{subfig}
\usepackage{todonotes}
\usepackage{rotating}

\usepackage{siunitx}
\newcommand{\timeout}{>\num{7.2e6}}
\usepackage{adjustbox}

\usepackage{newfloat}
\usepackage{listings}

\DeclareCaptionStyle{ruled}{labelfont=normalfont,labelsep=colon,strut=off} 
\floatstyle{ruled}
\newfloat{listing}{tb}{lst}{}
\floatname{listing}{Listing}
\usepackage{multirow}

\usepackage[switch, modulo]{lineno}

\usepackage{xcolor}

\title{MinePlanner: A Benchmark for Long-Horizon \\
Planning in Large Minecraft Worlds}
\author {
    William Hill\equalcontrib \textsuperscript{\rm 1},
    Ireton Liu\equalcontrib \textsuperscript{\rm 1},
    Anita De Mello Koch\textsuperscript{\rm 2},
    Damion Harvey\textsuperscript{\rm 1}, \\
    Nishanth Kumar\textsuperscript{\rm 3},
    George Konidaris\textsuperscript{\rm 2},
    Steven James\textsuperscript{\rm 1}
}
\affiliations {
    \textsuperscript{\rm 1}School of Computer Science and Applied Mathematics, University of the Witwatersrand\\
    \textsuperscript{\rm 2}Department of Computer Science, Brown University \\
    \textsuperscript{\rm 3}MIT Computer Science and Artificial Intelligence Laboratory 
}

\usepackage{bibentry}

\newcommand{\minecraft}{Minecraft\xspace}
\newcommand{\mineplanner}{MinePlanner\xspace}
\newcommand{\ntasks}{45\xspace}

\begin{document}

\maketitle

\begin{abstract}
We propose a new benchmark for planning tasks based on the Minecraft game. Our benchmark contains \ntasks tasks overall, but also provides support for creating both propositional and numeric instances of new Minecraft tasks automatically. We benchmark numeric and propositional planning systems on these tasks, with results demonstrating that state-of-the-art planners are currently incapable of dealing with many of the challenges advanced by our new benchmark, such as scaling to instances with thousands of objects. Based on these results, we identify areas of improvement for future planners.
Our framework is made available at \url{https://github.com/IretonLiu/mine-pddl/}.
\end{abstract}

\section{Introduction}

A major challenge in AI is the construction of autonomous agents capable of solving extremely long-horizon tasks. 
While approaches such as reinforcement learning (RL) struggle with such tasks, especially with sparse feedback, task-level planners are well-suited to such problems.
Additionally, these planners are typically domain-independent and so can be applied to a wide variety of problems, which is necessary if we desire generally intelligent agents.

However, these approaches require an abstract representation of a problem (typically using a structured language such as PDDL \citep{mcdermott1998pddl}) as input. 
Furthermore, these representations are carefully crafted by a human designer to contain only the necessary information required to solve the task \cite{fishman2020task}.
If we hope to scale these approaches to real-world tasks and develop truly autonomous agents, then planners must be capable of operating in domains that contain a large number of objects that may or may not be relevant to the task at hand.

While the issue of scaling to large domains is currently an area of active research \citep{illanes2019generalized}, current planning domains continue to focus on simplified world models by simply increasing the number of objects present in standard benchmarks \citep{silver2021planning}.
This, however, fails to accurately reflect the difficulty a planner would face in a noisy real-world task.
Additionally, recent work has demonstrated how PDDL representations can be directly learned from data  \citep{asai2018classical,james2020learning,ahmetoglu2022deepsym,silver2023inventing}. While these representations are often sound \citep{konidaris2018skills}, they typically contain many irrelevant symbols and action operators \citep{james2022autonomous}; planners that are robust to this issue would further bridge the gap between learning and planning. 

\begin{figure}[t]
\centering
\includegraphics[width=\linewidth]{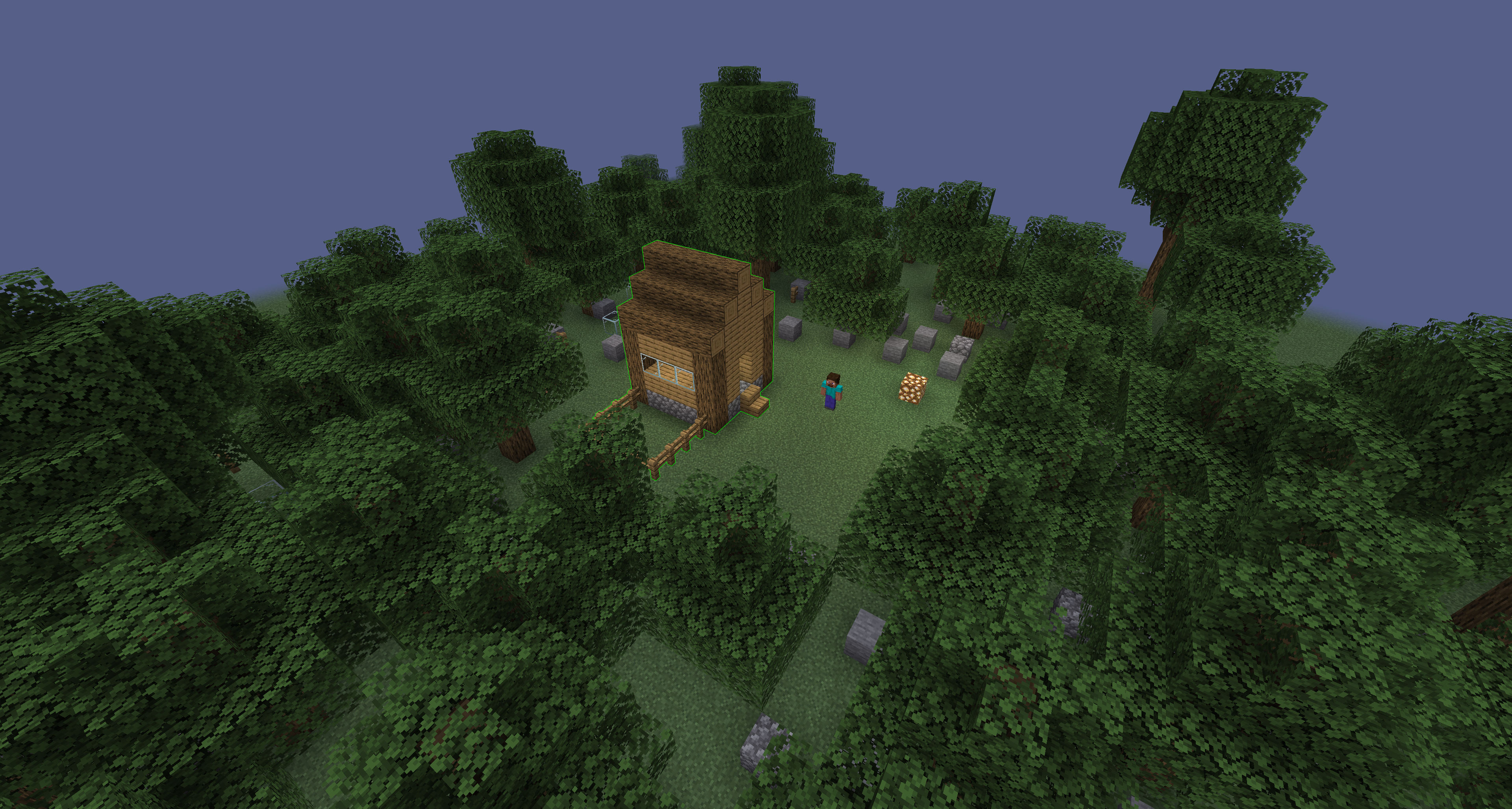}
\caption{A classical planning problem in our benchmark, requiring the agent to collect the necessary blocks and build a log cabin (outlined in green). This task contains over 5000 objects that the agent must reason about, including many that make up the surrounding blocks and ground that are irrelevant to the goal.}
\label{fig:cabin}
\end{figure}

Concurrently, the game of \minecraft has recently emerged as a promising testbed for RL research \citep{johnson2016malmo}, with its open-ended nature serving as a valuable proxy for the real world.
As a domain, \minecraft has several desirable characteristics for planning research: (a) it supports a wide variety of tasks in the form of structures that can be assembled, all of which require long-term planning (see Figure \ref{fig:cabin} for an example); (b) the game is naturally populated by objects in the form of blocks and items, removing the need for a human designer to inflate the number of objects in the domain artificially; and (c) there is an inherent hierarchy in building large \minecraft structures \citep{beukman2023hierarchically}, which may be of interest to hierarchical planners \citep{holler2020hddl}.


In this paper, we present \mineplanner, a long-horizon planning benchmark in large \minecraft worlds.
Our framework is capable of automatically generating tasks and verifying solutions in \minecraft, and supports both propositional and numeric planners.
We additionally provide a collection of \ntasks tasks which we used to benchmark representative propositional and numeric planners.
These results show that there is significant work still left to be done for planning in large domains with many objects.


\section{Related Work}

Minecraft has proven to be a popular testbed for machine learning research \cite{johnson2016malmo}.
One such platform is \textit{MineRL} \citep{guss2019minerl}, which provides a set of Minecraft-related tasks for the agent to solve.
More recently, \citet{fan2022minedojo} incorporate background knowledge of the game in the form of online documentation, forums and videos of human gameplay to assist agents in learning to play the game. 
In both cases, they typically require an agent to act in a high-dimensional environment with partially observable pixel input. 
These benchmarks require agents to grapple with multiple problems simultaneously---high-dimensional function approximation, continuous control, partial observability and long-term planning.
While a generally capable agent will ultimately be required to solve all of these problems, it also makes progress along any of these dimensions difficult to measure.
Our proposed benchmark abstracts away these low-level intricacies, allowing researchers to focus on what is perhaps the most interesting aspect of Minecraft---the ability to create impressive structures, such as entire cities \citep{salge2020ai}, through long-term planning.

\citet{aluru2015minecraft} provide a framework for converting small, constrained Minecraft problems into object-oriented MDPs where A* search could be applied. However, only relevant objects were specified, and blocks that constituted the boundary walls and ground were not explicitly represented. By contrast, we represent open-world settings as generically as possible. \citet{wichlacz2019construction} represent Minecraft construction tasks using PDDL and HTN formalisms; however, their representation abstracts much of the low-level complexity of Minecraft, most notably navigation, effectively treating the problem as a Blocks World. 
Finally, \citet{roberts2017automated} automatically generate Minecraft representations using various extensions of PDDL to handle open worlds and partial observability, but constrain the observable range to make planning feasible.  

One immediate challenge presented by Minecraft is the number of objects in the world. 
While previous work has identified the need to apply planners to domains with hundreds of objects, these domains are either scaled-up versions of classic problems such as \textit{Blocksworld} \cite{silver2021planning} or are created by combining multiple existing planning problems to introduce irrelevant objects \cite{fishman2020task}. 
While these testbeds may be useful for developing better planning algorithms, the domains are disjoint from those considered by the RL community, squandering the opportunity for collaboration between the fields.
Furthermore, these approaches may not capture the true complexity of the real world, which often contains objects and actions that may or may not be relevant to a given task. 

Finally, while our benchmark hopes to spur research in planning, it can be combined with tools such as PDDLGym \cite{silver2020pddlgym} to act as a reinforcement learning environment. This would benefit RL researchers who wish to focus on the challenging long-term planning problem posed by Minecraft, while avoiding the complexity of continuous control in pixel space.  


\section{A Framework for Generating \minecraft Planning Tasks}


We now present \mineplanner: a framework for generating \minecraft planning tasks that makes use of the APIs provided by MineDojo \citep{fan2022minedojo}. 
At the highest level, we define a specification schema for tasks that are used to generate \minecraft worlds.
We next extract objects and states (such as the agent's inventory) from the world and automatically generate a PDDL representation that can be used by planners.
To provide support for multiple approaches, the framework can generate both numeric and propositional PDDL. 
The difference between the two is primarily how locations are represented, and we discuss this further in subsequent sections. 

Our framework also supports the verification and visualisation of a plan---given the output of a planner, \mineplanner executes the proposed action in the game and verifies that the necessary predicates are achieved to solve the task. The frames collected during this process are saved and exported to video, which can then be used to promote research in a visually appealing manner. 
Finally, we provide a utility for extracting a list of objects, along with their coordinates, from saved \minecraft worlds\footnote{Using the PyBlock library (\url{github.com/alex4200/PyBlock}).} allowing users to easily specify new tasks without having to manually list the position of each object in the world. 


\subsection{Minecraft World Specification}

We define a task as a set of blocks and items that are initially placed in the Minecraft world and the agent's inventory. For simplicity, we restrict the items to only those that can be placed in the world (e.g. wood blocks, but not pickaxes). We specify the goal of the task as a set of blocks that are to be placed in the world at some location, a set of items the agent must have in its inventory, the agent's location, or any combination of the three. 
An example of a task specification is shown in Listing \ref{lst:yaml_task}, where an agent must place a log at location $(0, 4, 2)$ and additionally have at least one log remaining in its inventory to solve the task.

\begin{listing}[h]%
\caption{Example task specification in YAML.}%
\label{lst:yaml_task}%
\begin{lstlisting}[]
name: "Example Problem"
blocks:
  - position:
      x: '0'
      y: '4'
      z: '1'
    type: obsidian
items:
  - position:
      x: '1'
      y: '5'
      z: '5'
    quantity: 1
    type: diamond
inventory:
  - type: log
    quantity: '64'
  - type: obsidian
    quantity: '64
goal:
  agent:
    - position:
        x: '6'
        y: '4'
        z: '-5'
  blocks:
    - position:
        x: '0'
        y: '4'
        z: '-2'
      type: log
  inventory:
    - type: log
      quantity: '1'
\end{lstlisting}
\end{listing}

To produce a tractable representation of a \minecraft world, we make the following simplifications that vary slightly from the original game:

\begin{itemize}
    \item Each task is created using a flat world with a single layer of grass blocks serving as the ground.
    \item There are no non-player characters, and items placed in the world do not despawn.\footnote{Both of these would violate the frame assumption \citep{pasula2004learning}.}
    \item The agent does not require the necessary tools to break certain blocks. For example, the agent can break a tree block without an axe.
    \item Broken blocks are immediately added to the agent's inventory without being dropped on the ground as items. 
    \item The agent is constrained by allowing it to move only one unit (block) in any cardinal direction at a given time.
\end{itemize} 

\subsection{State Representations}

The types of each object, such as \texttt{agent}, \texttt{grass-block} or \texttt{flower}, are specified directly by \minecraft itself.
To represent the state of the world, we must keep track of the location of all items and objects, as well as the agent's inventory.
Using PDDL 2.1 \citep{fox2003pddl2}, this is relatively straightforward: the domain is defined by a predicate governing whether an object is present in the world (or in the agent's inventory), or whether it has been destroyed, and several numeric fluents that keep track of each object's  $x$, $y$ and $z$ positions. Fluents are also used to track how many items are in the agent's inventory, since the agent can collect multiple objects of the same type. There is one fluent for each type present in the world; for example, to track the number of flowers in the inventory, we would have the following: \texttt{(agent-num-flower ?ag - agent)}.

For planners that do not support numeric fluents, we represent positions and count using predicates only. This is achieved by defining ``integer'' objects (e.g., \texttt{position36}, \texttt{count0}) along with predicates that enforce relationships between these objects, such as sequentiality (\texttt{(are-seq ?x1 - int ?x2 - int)}) and whether an object is at a particular location (e.g.,  \texttt{(at-x ?l - locatable ?x - position)}). The inventory is represented by determining whether the count of a particular item matches some number.\footnote{\minecraft enforces a maximum inventory count of 64 per object, which can be enumerated.} The predicate equivalent to its numeric counterpart is \texttt{(agent-has-n-flower ?ag - agent ?n - count)}.


\subsection{Operator Representations}



We model two types of operators in our \minecraft worlds: \textit{movement} and \textit{interaction} actions. 
Movement involves the agent navigating one unit in a cardinal direction, and includes the ability to jump in a particular direction. As in the game, an agent's movement is restricted by objects around it, and the preconditions for movement operators reflect this. 

Another challenge in \minecraft is that there is no explicit action for picking up an item---an agent simply walks over an item to collect it. 
To avoid inconsistency between the PDDL representations and the game, we account for this by introducing two separate actions for every movement: a movement action that \textit{cannot} be executed if the destination is occupied by an item and an action that combines a movement with a pickup. Additionally, the agent can only move to a destination above an existing block. The complete list of movement operators is as follows:




\begin{itemize}
    \item \texttt{move\_[direction]}: Moves the agent by one block in the specified direction. The agent cannot move to a location that is occupied by an item. An example of moving north without collecting an object is given by Listing~\ref{lst:pddl_num_movement}.
    \item \texttt{move\_and\_pickup\_[item]\_[direction]}: Moves the agent by one block in the specified direction and collects an item at the resulting position.

    \item \texttt{jumpup\_[direction]}: Moves the agent by one block in the specified direction and one block along the positive vertical axis. The agent cannot move into a position that is occupied by an item. 
    \item \texttt{jumpdown\_[direction]}: Moves the agent by one block in the specified direction and one block along the negative vertical axis. The agent cannot move into a position that is occupied by an item. 
    
    \item \texttt{jumpup\_and\_pickup\_[item]\_[direction]}: Moves the agent by one block in the specified direction and one block along the positive vertical axis and collects an item at the resulting position.
    \item \texttt{jumpdown\_and\_pickup\_[item]\_[direction]}: Moves the agent by one block in the specified direction and one block along the negative vertical axis and collects an item at the resulting position.
\end{itemize}


\begin{listing}[h]%
\caption{Example of a movement action using fluents}%
\label{lst:pddl_num_movement}%
\begin{lstlisting}[language=Lisp]
(:action move-north
 :parameters (?ag - agent)
 :precondition (and (agent-alive ?ag) 
  (exists (?b - block) (and 
  (block-present ?b) (= (x ?b) (x ?ag)) 
  (= (y ?b) (+ (y ?ag) -1)) (= (z ?b) 
  (+ (z ?ag) -1)))) (and 
  (not (exists (?b - block) (and 
  (block-present ?b) (= (x ?b) (x ?ag)) 
  (or (= (y ?b) (+ (y ?ag) 1)) (= (y ?b) (y ?ag))) 
  (= (z ?b) (+ (z ?ag) -1))))) 
  (not (exists (?i - item) (and (item-present ?i) (= (x ?i) (x ?ag)) (= (y ?i) (y ?ag)) (= (z ?i) 
  (+ (z ?ag) -1)))))))
 :effect (and (decrease (z ?ag) 1))
)\end{lstlisting}
\end{listing}

The agent is also capable of manipulating blocks, and we support two such interaction operators:
\begin{itemize}
    \item \texttt{place\_[block]\_[direction]}: Places a block from the agent's inventory one block in front of the agent in the specified direction. The agent cannot place a block in a position that is occupied by another block or item and there must be a block below the target location.
    
    \item \texttt{break\_[block]\_[direction]}: Breaks the block one unit in front of the agent in the specified direction. The block is collected and added to the agent's inventory. The agent cannot break a block that has an item on top of it --- in such a case, the agent will first have to pick the item up and then break the block.  An example of breaking a grass block is given by Listing~\ref{lst:pddl_prop_inter}.
\end{itemize}


\begin{listing}[!]%
\caption{Example of an interaction action using predicates}%
\label{lst:pddl_prop_inter}%
\begin{lstlisting}[language=Lisp]
(:action break-grass_block-north
 :parameters (?ag - agent ?b - grass_block-block ?x - position ?y - position ?y_up - position ?z - position ?z_front - 
  position ?n_start - count ?n_end - count)
 :precondition (and (agent-alive ?ag)  (at-x ?ag ?x)  (at-y ?ag ?y) (at-z ?ag ?z)  (at-x ?b ?x) (at-y ?b ?y) (at-z ?b ?z_front) (are-seq ?z_front ?z) (are-seq ?y ?y_up)
  (block-present ?b) (not (exists (?i - item) (and (item-present ?i)
  (at-x ?i ?x) (at-y ?i ?y_up) (at-z ?i ?z_front)))) (are-seq ?n_start ?n_end) (agent-has-n-grass_block ?ag ?n_start)
  )
 :effect (and (not (block-present ?b)) 
  (not (at-x ?b ?x)) (not (at-y ?b ?y)) (not (at-z ?b ?z_front))
  (not (agent-has-n-grass_block ?ag ?n_start)) 
  (agent-has-n-grass_block ?ag ?n_end))
)
\end{lstlisting}
\end{listing}

\subsection{Goal Representations}

Since \minecraft objects of the same type are interchangeable, we use existential preconditions when specifying a goal related to the location of a block. This allows us to specify that, for example, \textit{any} wood block should be placed at a particular location, since all wood blocks are functionally identical. However, not all planners provide support for the \texttt{exist} keyword in the goal specification. 
To make our representation as accessible as possible, we introduce a ``virtual'' operator called \texttt{checkgoal} whose precondition is the actual condition for solving the task and whose effect sets a predicate \texttt{goal-achieved} to true. 
This is the only operator that can affect \texttt{goal-achieved}, which allows us to specify that the goal for all tasks is simply  \texttt{goal-achieved}.
For numeric planning, an example of a task whose goal is to place a \texttt{planks-block} at location $(0,4,2)$ is given by Listing \ref{lst:goal_num}, while Listing \ref{lst:goal_prop} shows the corresponding propositional equivalent.

\begin{listing}[h]%
\caption{Example of goal attainment using fluents}%
\label{lst:goal_num}%
\begin{lstlisting}[language=Lisp]
(:action checkgoal
 :parameters (?ag - agent)
 :precondition (and (agent-alive ?ag)  
  (exists (?b - planks-block) 
  (and (block-present ?b) (= (x ?b) 0)
  (= (y ?b) 4)(= (z ?b) 2))))
 :effect (and (goal-achieved ?ag))
)
\end{lstlisting}
\end{listing}

\begin{listing}[h!]%
\caption{Example of goal attainment using predicates}%
\label{lst:goal_prop}%
\begin{lstlisting}[language=Lisp]
(:action checkgoal
 :parameters (?ag - agent)
 :precondition (and (agent-alive ?ag)
  (exists (?b - planks-block) (and 
  (block-present ?b) (at-x ?b position0)
  (at-y ?b position4) (at-z ?b position2)))
  )
 :effect (and (goal-achieved ?ag))
)
\end{lstlisting}
\end{listing}
    
\subsection{Generating Solutions}

Since we expect our tasks to be beyond the capabilities of current planners, we introduce a modification to Minecraft that allows human experts to play any of the defined tasks and generate a plan from their actions.
The action trace of the player (consisting of player movements and interactions with blocks) is logged to a file that is then parsed into a format suitable for verification in \mineplanner. 
This subsystem serves two purposes: (a) it allows us to easily produce at least one satisficing plan, even in extremely complex tasks, and (b) it serves as a human benchmark against which the solutions produced by planners can be compared.

\begin{table*}[h!]
    \centering
    \renewcommand{\arraystretch}{0.9}
    \begin{tabular}{cccccccl}
        \toprule
       Task & Variant & \makecell{Observation \\ Range} &  \makecell{Inital \\ Objects}  &  \makecell{Initial\\ Predicates}  &  \makecell{Goal \\ Pred.}  & 
       \makecell{Human Sol. \\ Length}  & Description \\
        \midrule
        \multirow{3}{*}{\texttt{move}} & Easy & $(13, 9, 13)$ & $0$ & $762/851$ & $1$ & $5$ & \multirow{3}{4cm}{Navigate to a specific location.}\\
         & Medium & $(21, 15, 21)$ & $12$ & $1908/2273$ & $1$ & $11$ & \\
         & Hard   & $(71, 31, 71)$ & $1071$ & $23706/29432$ & $1$ & $25$ & \\
        \midrule
        \multirow{3}{*}{\makecell{\texttt{pickup}\\\texttt{diamond}}} & Easy & $(13, 9, 13)$ & $2$ & $767/856$ & $1$ & $6$&  \multirow{3}{4cm}{Navigate and pickup a single diamond in the world.} \\
         & Medium & $(21, 15, 21)$ & $8$ & $1893/2254$ & $1$ & $7$& \\
         & Hard   & $(71, 31, 71)$ & $1072$ & $43870/54642$ & $1$ & $23$& \\
              \midrule
                 \multirow{3}{*}{\makecell{\texttt{gather}\\\texttt{wood}}} & Easy & $(13, 9, 13)$ & $1$ & $767/857$ & $1$ & $3$&  \multirow{3}{4cm}{Navigate and pickup a single log in the world.}\\
         & Medium & $(21, 15, 21)$ & $6$ & $1881/2240$ & $1$ & $4$& \\
         & Hard   & $(71, 31, 71)$ & $1071$ & $43859/54637$ & $1$ & $5$& \\
        \midrule
                \multirow{3}{*}{\makecell{\texttt{place}\\\texttt{wood}}} & Easy & $(13, 9, 13)$ & $1$ & $767/856$ & $2$ & $3$&  \multirow{3}{4cm}{Navigate to a specific location and place a log from inventory.}\\
         & Medium & $(21, 15, 21)$ & $13$ & $1917/2284$ & $2$ & $11$& \\
         & Hard   & $(71, 31, 71)$ & $1071$ & $43863/54626$ & $2$ & $25$& \\
        \midrule
                \multirow{3}{*}{\makecell{\texttt{pickup}\\\texttt{and}\\\texttt{place}}} & Easy & $(13, 9, 13)$ & $1$ & $767/857$ & $1$ & $16$&  \multirow{3}{4cm}{First locate a plank in the world, then place it at a specific location.}\\
         & Medium & $(21, 15, 21)$ & $16$ & $1926/2296$ & $1$ & $18$& \\
         & Hard   & $(71, 31, 71)$ & $1071$ & $43889/29432$ & $1$ & $42$& \\
        \midrule
                \multirow{3}{*}{\makecell{\texttt{gather}\\\texttt{multi}\\\texttt{wood}}} & Easy & $(13, 9, 13)$ & $3$ & $775/867$ & $1$ & $7$&  \multirow{3}{4cm}{Navigate and pickup a three logs in the world.}\\
         & Medium & $(21, 15, 21)$ & $18$ & $1934/2306$ & $1$ & $20$& \\
         & Hard   & $(71, 31, 71)$ & $1071$ & $43865/54627$ & $1$ & $8$& \\
        \midrule
                \multirow{3}{*}{\texttt{climb}} & Easy & $(13, 9, 13)$ & $18$ & $825/928$ & $1$& $7$&  \multirow{3}{4cm}{Place a block at an elevated $y$ location by climbing a staircase.}\\
         & Medium & $(21, 15, 21)$ & $7$ & $1892/2253$ & $1$& $39$ & \\
         & Hard   & $(71, 31, 71)$ & $18$ & $20387/25308$ & $1$ & $105$& \\
        \midrule
                \multirow{3}{*}{\makecell{\texttt{cut}\\\texttt{tree}}} & Easy & $(21, 31, 21)$ & $60$ & $2113/2518$ & $1$ & $77$&  \multirow{3}{4cm}{Cut down a tree by removing its wood.}\\
         & Medium & $(41, 31, 41)$ & $75$ & $7144/8793$ & $1$& $295$ & \\
         & Hard   & $(65, 31, 65)$ & $946$ & $37080/46137$ & $9$ & $392$& \\
        \midrule
                \multirow{3}{*}{\makecell{\texttt{build}\\\texttt{bridge}}} & Easy & $(13, 9, 13)$ & $80$ & $773/862$ & $2$& $7$ &  \multirow{3}{4cm}{Build a wooden bridge over water.}\\
         & Medium & $(21, 15, 21)$ & $255$ & $1924/2294$ & $4$ & $56$& \\
         & Hard   & $(71, 31, 71)$ & $409$ & $20350/25263$ & $6$ & $65$& \\
        \midrule
                \multirow{3}{*}{\makecell{\texttt{build}\\\texttt{cross}}} & Easy & $(13, 9, 13)$ & $5$ & $788/882$ & $5$& $45$ &  \multirow{3}{4cm}{Collect blocks to build a cross shape.}\\
         & Medium & $(21, 15, 21)$ & $10$ & $1900/2264$ & $5$ & $84$& \\
         & Hard   & $(71, 31, 71)$ & $1071$ & $43854/54627$ & $5$ & $103$& \\
        \midrule
                \multirow{3}{*}{\makecell{\texttt{build}\\\texttt{wall}}} & Easy & $(13, 9, 13)$ & $9$ & $806/904$ & $9$ & $117$&  \multirow{3}{4cm}{Collect blocks to build a wall.}\\
         & Medium & $(21, 15, 21)$ & $16$ & $1927/2298$ & $9$ & $99$& \\
         & Hard   & $(71, 31, 71)$ & $1071$ & $43869/54627$ & $9$ & $102$& \\
        \midrule
                \multirow{3}{*}{\makecell{\texttt{build}\\\texttt{well}}} & Easy & $(13, 9, 13)$ & $26$ & $876/990$ & $26$ & $226$&  \multirow{3}{4cm}{Collect blocks to build a well.}\\
         & Medium & $(21, 15, 21)$ & $36$ & $2010/2400$ & $26$ & $443$& \\
         & Hard   & $(71, 31, 71)$ & $1071$ & $43870/54633$ & $26$ & $420$& \\
      \midrule
                \multirow{3}{*}{\makecell{\texttt{build}\\\texttt{shape}}} & Easy & $(13, 9, 13)$ & $1$ & $772/861$ & $5$ & $31$&  \multirow{3}{4cm}{Build a variety of shapes with items from inventory.}\\
         & Medium & $(21, 15, 21)$ & $12$ & $1923/2288$ & $9$ & $74$& \\
         & Hard   & $(71, 31, 71)$ & $1071$ & $43871/54663$ & $27$ & $138$& \\
      \midrule
                \multirow{3}{*}{\makecell{\texttt{collect}\\\texttt{and build}\\\texttt{shape}}} & Easy & $(13, 9, 13)$ & $5$ & $788/882$ & $5$ & $40$&  \multirow{3}{4cm}{Collect blocks to build a variety of shapes.}\\
         & Medium & $(21, 15, 21)$ & $11$ & $1910/2275$ & $11$ & $127$& \\
         & Hard   & $(71, 31, 71)$ & $1071$ & $43864/54647$ & $27$ & $292$& \\
        \midrule
                \multirow{3}{*}{\makecell{\texttt{build}\\\texttt{cabin}}} & Easy & $(21, 11, 21)$ & $0$ & $1892/2246$ & $116$ & $481$&  \multirow{3}{4cm}{Build a log cabin.}\\
         & Medium & $(41, 11, 41)$ & $116$ & $7318/9008$ & $116$ & $904$& \\
         & Hard   & $(65, 11, 65)$ & $5019$ & $18201/22583$ & $116$ & $1268$& \\
        \bottomrule
    \end{tabular}
    \caption{A list of tasks provided by \mineplanner and their relevant statistics. 
    Initial Objects does not include the ground (i.e. only objects explicitly specified in the YAML configuration are included). Initial Predicates is the number of predicates specified in the initial state of the problem file and is formatted as proposition/numerical. Goal Predicates is the number of goal conditions specified in the YAML configuration. Human Solution Length refers to the length of a plan constructed from a human expert's playthrough when solving the task.}
    \label{tab:tasks}
\end{table*}

\section{Benchmark Tasks in \minecraft}

We create an initial suite of tasks, listed in Table~\ref{tab:tasks},  to serve as challenging problems for current planners.
To create a finite representation of a (near infinite) \minecraft world, we consider only those blocks within some radius of the agent's initial location, termed the \textit{observation range}.\footnote{The agent is centred in the \textit{observation range}.}

\begin{figure}[h!]
    \centering
    \subfloat[Easy variant]{
        \includegraphics[width=0.9\linewidth]{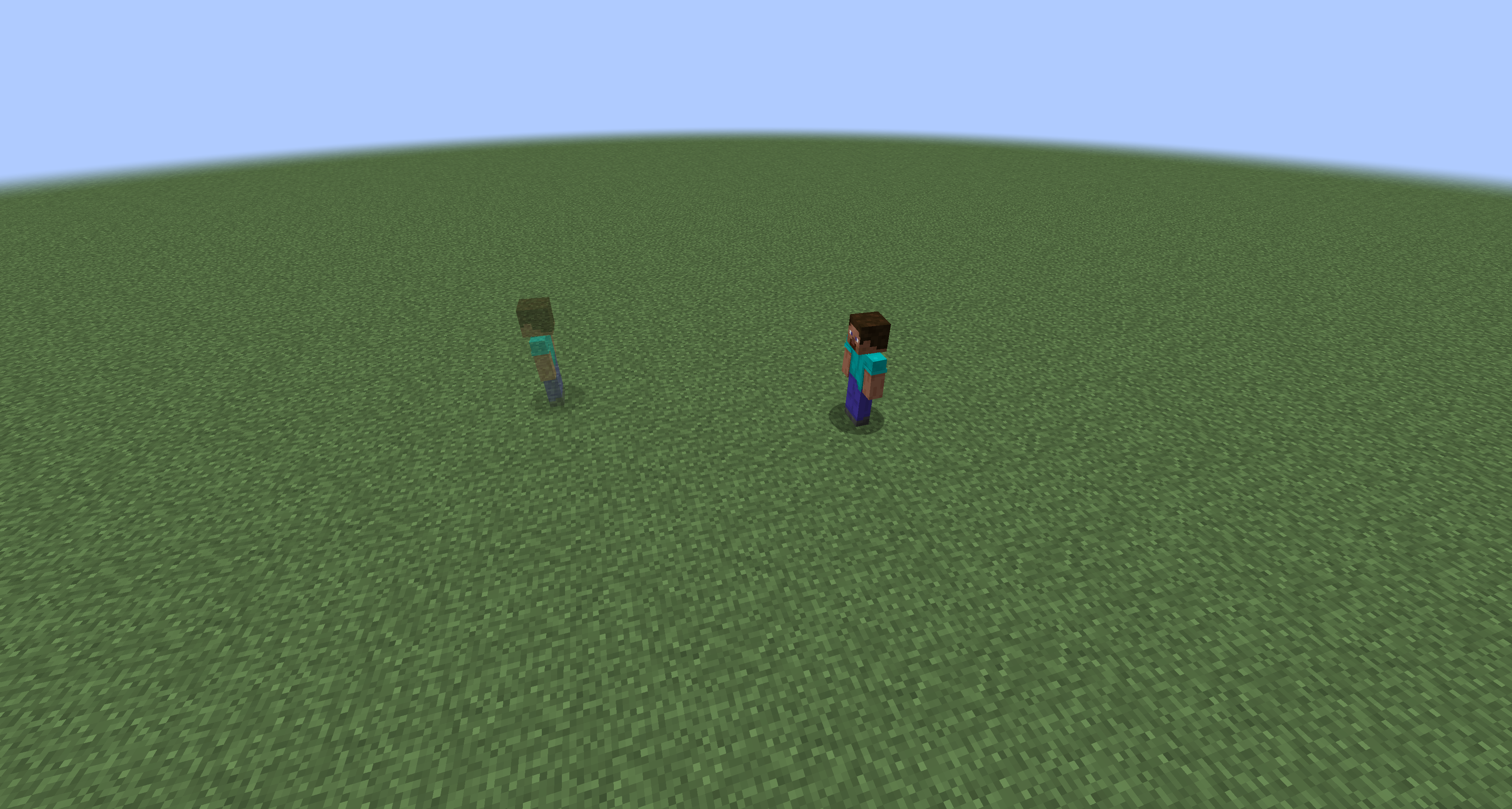}
        \label{fig:subfig1}
    }
    \\
    \subfloat[Medium variant]{
        \includegraphics[width=0.9\linewidth]{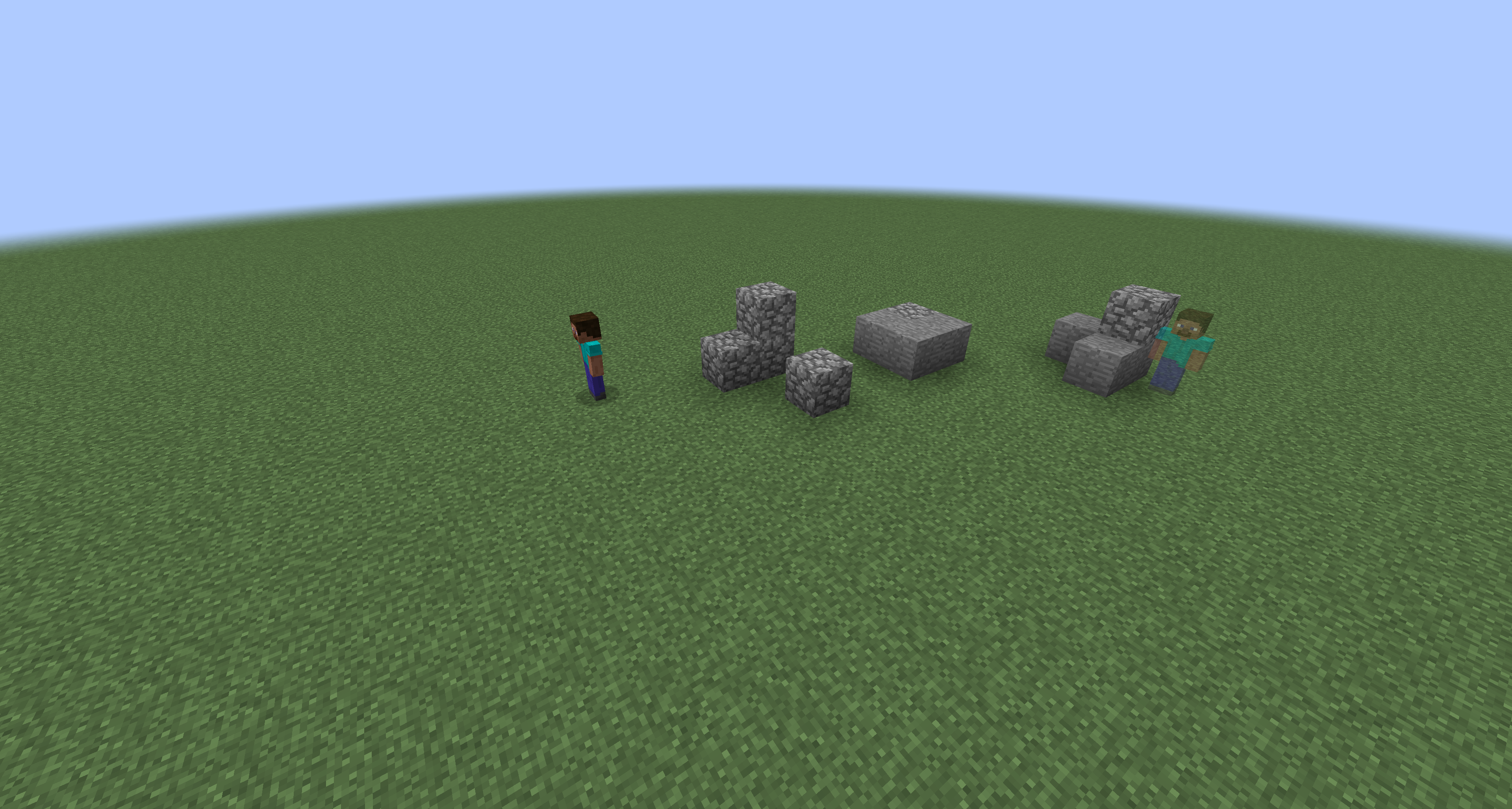}
        \label{fig:subfig2}
    }
    \\ 
    \subfloat[Hard variant]{
        \includegraphics[width=0.9\linewidth]{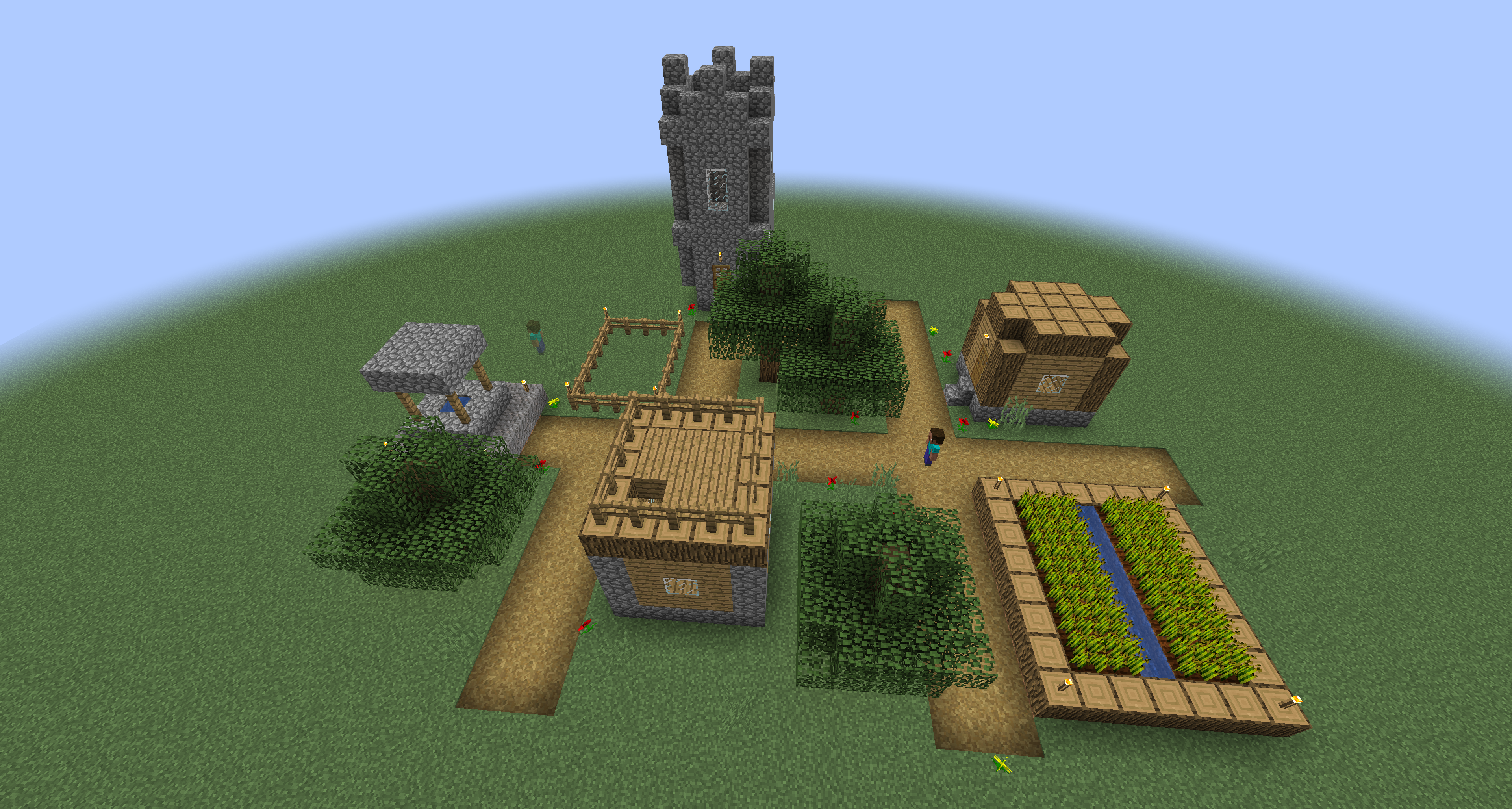}
        \label{fig:subfig3}
    }
    \caption{Three variants for the task of navigating to a particular location. (a) The easy task contains no irrelevant blocks, and so the world is empty. (b) The medium contains a few additional blocks which serve as obstacles and make navigation more challenging. (c) The hard task requires navigating within a small village consisting of hundreds of objects that are irrelevant for this particular task.}
    \label{fig:difficulties}
\end{figure}

We define 15 types of tasks, where each task has three difficulties. Broadly, the easiest version of a task contains only those blocks necessary to solve the task and also has the smallest observation range. A medium difficulty task contains more blocks that typically exist in the world, but are not strictly relevant to the task at hand. Finally, the hardest version of each task takes place in a ``realistic'' \minecraft setting with a much larger observation range.
The difference between difficulties is illustrated by Figure~\ref{fig:difficulties}.

\section{Experiments}




We first benchmark Fast Downward \citep{helmert2006fast}, a propositional planner with the LAMA configuration, and ENHSP-20 \citep{scala2020subgoaling}, a numerical planner with A* search and the AIBR heuristic \citep{scala2016interval}, on the tasks provided by \mineplanner.
All experiments were conducted using Apptainer \citep{singularity} on a cluster of AMD Ryzen 9 7900X3D CPUs, using 128 virtual cores and 250GB of RAM per trial.
We record planning time, including the amount of time spent preprocessing the PDDL file by each planner. For Fast Downward, this refers to the time taken to translate PDDL to SAS, while for ENHSP, this measures grounding.  We set a timeout limit of two hours for each task, since an autonomous agent must ultimately be capable of planning in large environments within a reasonable timeframe.
Results are reported in Table~\ref{tab:results}, with the means over five runs reported.\footnote{For readability, we include only the means here, but report the full table along with standard deviations in the appendix.}

\begin{table*}
\renewcommand{\arraystretch}{0.8}
\centering
    \begin{tabular}{cccccccc}
       \toprule
        \multirow{2}{*}{Task} & \multirow{2}{*}{Variant} & \multicolumn{3}{c}{FastDownward} & \multicolumn{3}{c}{ENHSP} \\
        \cmidrule(lr){3-5} \cmidrule(lr){6-8} 
         & & Transl. (s) & Search (ms) & Total (s) & Ground (s) & Search (ms) & Total (s)\\
        \midrule
        \multirow{3}{*}{\texttt{move}} & Easy & $29.55$ & $9.05$ & $29.80$ &  $9.75$ & $21.60$ & $\mathbf{25.19}$ \\
         & Medium & --- & --- &  --- & $49.05$ & $\timeout$ & $\timeout$ \\
         & Hard  & --- & --- & --- & --- & --- & --- \\
        \midrule
        \multirow{3}{*}{\makecell{\texttt{pickup}\\\texttt{diamond}}} & Easy & $242.85$ & $172.88$ & $\mathbf{246.46}$ &  $9.28$ & $\num{7.15e6}$ & $\num{7.19e6}$ \\
         & Medium & --- & --- & --- & $48.18$ & $\timeout$ & $\timeout$ \\
         & Hard  & --- & --- & --- & --- & --- & --- \\
        \midrule
        \multirow{3}{*}{\makecell{\texttt{gather}\\\texttt{wood}}} & Easy & $248.96$ & $511.03$ & $\mathbf{251.35}$ &  $7.42$ & $4\,352.49$ & $4\,382.79$ \\
         & Medium &--- & --- &  --- & $47.32$ & $\timeout$ & $\timeout$ \\
         & Hard  & --- & --- & --- & --- & --- & --- \\
        \midrule
        \multirow{3}{*}{\makecell{\texttt{place}\\\texttt{wood}}} & Easy & $608.85$ & $85420$ & $\mathbf{973.33}$ &  $7.53$ & $\timeout$ & $\timeout$ \\
         & Medium & --- & --- &  --- & $49.83$ & $\timeout$ & $\timeout$ \\
         & Hard  & --- & --- & --- & --- & --- & --- \\
        \midrule
        \multirow{3}{*}{\makecell{\texttt{pickup and}\\\texttt{place}}} & Easy &  $597.39$ & $525235$ &  $\mathbf{1134.46}$ & $7.36$ & $\timeout$ & $\timeout$ \\
         & Medium & --- & --- &  --- & $50.23$ & $\timeout$ & $\timeout$ \\
         & Hard  & --- & --- & --- & --- & --- & --- \\
        \midrule
        \multirow{3}{*}{\makecell{\texttt{gather multi}\\\texttt{wood}}} & Easy & --- & --- & --- &  $7.62$ & $\timeout$ & $\timeout$ \\
         & Medium & --- & --- &  --- & $50.19$ & $\timeout$ & $\timeout$ \\
         & Hard  & --- & --- & --- & --- & --- & --- \\
        \midrule
        \multirow{3}{*}{\makecell{\texttt{climb}}} & Easy & --- & --- & --- &  $8.70$ & $\timeout$ & $\timeout$ \\
         & Medium & --- & --- &  --- & $48.71$ & $\timeout$ & $\timeout$ \\
         & Hard  & --- & --- & --- & --- & --- & --- \\
        \midrule
        \multirow{3}{*}{\makecell{\texttt{cut}\\\texttt{tree}}} & Easy & --- & --- & --- &  $60.35$ & $\timeout$ & $\timeout$ \\
         & Medium & --- & --- &  --- & --- & --- & --- \\
         & Hard  & --- & --- & --- & --- & --- & --- \\
        \midrule
        \multirow{3}{*}{\makecell{\texttt{build}\\\texttt{bridge}}} & Easy & --- & --- & --- &  $13.38$ & $\timeout$ & $\timeout$ \\
         & Medium & --- & --- &  --- & $57.10$ & $\timeout$ & $\timeout$ \\
         & Hard  & --- & --- & --- & --- & --- & --- \\
        \midrule
        \multirow{3}{*}{\makecell{\texttt{build}\\\texttt{cross}}} & Easy & --- & --- & --- &  $8.31$ & $\timeout$ & $\timeout$ \\
         & Medium & --- & --- &  --- & $48.34$ & $\timeout$ & $\timeout$ \\
         & Hard  & --- & --- & --- & --- & --- & --- \\
        \midrule
        \multirow{3}{*}{\makecell{\texttt{build}\\\texttt{wall}}} & Easy & --- & --- & --- &  $8.48$ & $\timeout$ & $\timeout$ \\
         & Medium & --- & --- &  --- & $50.43$ & $\timeout$ & $\timeout$ \\
         & Hard  & --- & --- & --- & --- & --- & --- \\
        \midrule
        \pagebreak
        \multirow{3}{*}{\makecell{\texttt{build}\\\texttt{well}}} & Easy & --- & --- & --- &  $10.74$ & $\timeout$ & $\timeout$ \\
         & Medium & --- & --- &  --- & $54.40$ & $\timeout$ & $\timeout$ \\
         & Hard  & --- & --- & --- & --- & --- & --- \\
        \midrule
        \multirow{3}{*}{\makecell{\texttt{build}\\\texttt{shape}}} & Easy & --- & --- & --- &  $8.06$ & $\timeout$ & $\timeout$ \\
         & Medium & --- & --- &  --- & $51.70$ & $\timeout$ & $\timeout$ \\
         & Hard  & --- & --- & --- & --- & --- & --- \\
        \midrule
        \multirow{3}{*}{\makecell{\texttt{collect and}\\\texttt{build shape}}} & Easy & --- & --- & --- &  $8.19$ & $\timeout$ & $\timeout$ \\
         & Medium & --- & --- &  --- & $51.10$ & $\timeout$ & $\timeout$ \\
         & Hard  & --- & --- & --- & --- & --- & --- \\
        \midrule
        \multirow{3}{*}{\makecell{\texttt{build}\\\texttt{cabin}}} & Easy & --- & --- & --- &  ---& ---& ---\\
         & Medium & --- & --- &  --- & ---& ---& ---\\
         & Hard  & --- & --- & --- & --- & --- & --- \\
        \bottomrule
    \end{tabular}
        \caption{The running times for Fast Downward and ENHSP-20 when run on the MinePlanner task suite. No result could be obtained for results marked as --- because the planner failed to translate (in the case of Fast Downward) or ground (for ENHSP-20). Entries marked $\timeout$ indicate that the planner timed out.}
        \label{tab:results}
\end{table*}

\subsection{Domain-Independent Planning Results}

\begin{figure}[h!]
    \centering
    \subfloat[Translation time for increasing world size.]{
        \includegraphics[width=0.85\linewidth]{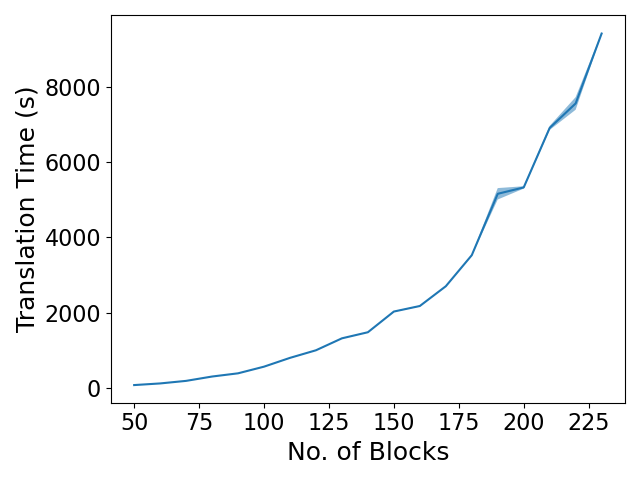}
    }
    \\
    \centering
    \subfloat[Planner search time for increasing world.size]{
        \includegraphics[width=0.85\linewidth]{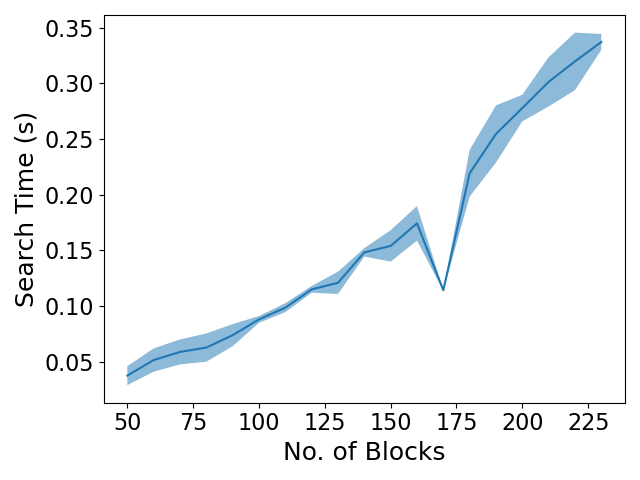}
    }
    \caption{The translation and planner search time is shown for the \texttt{move} task, starting from the size of the easy variant, and increasing until the FastDownward Planner can no longer translate the problem. The solid line and shaded areas represent the mean and standard deviation over five runs.}
    \label{fig:scaled_plots}
\end{figure}

The results indicate that the majority of tasks could not be solved by either planner. 
The translation step of Fast Downward was particularly problematic, with most of the tasks exhausting all memory before the file was translated to SAS. However, for those tasks where translation was successful, the subsequent search procedure was extremely fast (taking a few seconds at most). 
To investigate this behaviour, we conduct a further experiment using the \texttt{move} task. We begin with the Easy variant, and then incrementally scale the size of the world until Fast Downward fails to translate.
These results, shown in Figure \ref{fig:scaled_plots}, how translation time scales exponentially as the size of the world increases.
This is a worrying trend, despite the linear search time, as it indicates that Fast Downward does not scale well with the number of objects in the world.

By contrast, ENHSP-20 is capable of grounding almost all of the problems, but fails to find a successful plan for most. 
For tasks that were successfully solved, planning is significantly slower than Fast Downward, illustrating the tradeoff between the two.

\subsection{Automatically Removing Irrelevant Objects}

Given the issues caused by the number of objects in the world, it is natural to wonder whether many of these objects are irrelevant and can be pruned. 
We therefore apply \textit{task scoping}, a recently introduced algorithm for preprocessing a PDDL file to remove provably irrelevant objects and operators, which has been shown to produce significant speedups in classic problems \citep{fishman2020task}.
We apply this preprocessing step on the numeric domains that failed to plan in a reasonable amount of time. We select these tasks, since (a) task scoping requires the problem to first be preprocessed, so any tasks that could not complete the grounding step cannot be scoped using this method, and (b) the removal of irrelevant objects should improve planner performance.

Unfortunately, our results indicate that the scoped environments have no effect on the final run time of the planner.
Table \ref{tab:scope_results} reports the average number of objects, actions and grounded actions removed across all tasks.
In particular, task scoping fails to realise any advantages in the easy and medium task variants, since it considers the entire domain to be relevant and only removes grounded actions. 
We might expect to see gains for the hard variants, as they include many irrelevant objects. However, because the algorithm first requires that the problem be grounded, task scoping cannot be applied to these problems.

\begin{table}[h!]
    \centering
    \begin{tabular}{ccc}
    \toprule
        \makecell{Actions \\ Removed} & \makecell{Objects \\ Removed} & \makecell{Grounded Actions \\ Removed}  \\ \midrule
        $0\pm{0}$ & $0\pm{0}$ & $1009.25\pm{426}$ \\
        \bottomrule
    \end{tabular}
    \caption{The number of actions, objects and grounded actions removed by the task scoping algorithm. Results are averaged across all tasks in which the numeric planner failed to plan in a reasonable amount of time, with mean and standard deviation reported. }
    \label{tab:scope_results}
\end{table}

\subsection{Lifted Planning}
Since grounding tasks in \mineplanner has shown to be problematic, we also investigate whether planners that operate directly on the lifted representation could prove to be a solution.
To this end, we attempt to run the Powerlifted planning system \citep{correa2020lifted} in its recommended satisficing configuration on our problems.\footnote{We also attempted to run QPlanner \citep{shaik2021classical}, but it was unable to parse and begin searching on even our easiest tasks.} This requires creating modified versions of all the operators in our domains to remove negations and existentials from the preconditions.

We found that Powerlifted quickly exhausts all 250GB of RAM during search on all of the easy variants of our tasks. As such, the planner is unable to find a solution to any of our tasks.
This finding suggests that our tasks present significant challenges beyond existing benchmark problems for lifted planning (e.g. \citet{ijcai2021p567}), which Powerlifted is able to largely solve. Studying the source of the significant memory requirements for lifted planning in our problems could provide useful directions for future planning research.

\section{Conclusion}

We presented \mineplanner---a framework for generating \minecraft tasks in PDDL that can serve as challenging domains for classical planners. We also proposed a set of 45 initial tasks, varying in difficulty, and benchmarked two domain-independent planners on these domains. 
The results indicated that there is still a significant technical gap to overcome to solve these large problems, with no planner capable of solving any of the hard tasks.

One future direction is to leverage modern computing clusters to solve these challenging tasks, as has been done for previous ``grand challenges'' \citep{silver2016mastering}. 
However, at present, there are several issues that prevent the full utilisation of our hardware. 
While planners like ENHSP-20 could potentially benefit from parallelisation, especially due to their long search times, they are also not able to ground larger problems despite being provided with 250GB of RAM.
Similarly, the tasks Fast Downward failed to complete were due to the high memory requirements needed during the translation phase.
Modern planners will require more than parallelisation to scale effectively to object-dense environments.

If we wish to apply planning in realistic domains, we require planners that can operate within these object-dense environments in a reasonable amount of time, using a reasonable amount of memory.
We hope that our benchmark will serve as a catalyst for developing new approaches to planning in complex domains, and form a bridge between the learning and planning communities.

\clearpage

\bibliography{aaai24}

\onecolumn
\newpage

\appendix

\section*{Appendix}

\subsection*{A. Full Results}

In Table~\ref{tab:results}, we benchmarked Fast Downward and ENHSP-20 on the tasks provided with \mineplanner.
Table~\ref{tab:full-results} below extends these results by reporting both the means and standard deviations over 5 runs.  All experiments were conducted using Apptainer \citep{singularity} on a cluster of AMD Ryzen 9 7900X3D CPUs, using 128 virtual cores and 250GB of RAM.

\renewcommand{\arraystretch}{0.4}
\begin{table}[h!]
\centering
\rotatebox{90}{ 
\resizebox{1.03\textwidth}{!}{
\begin{tabular}{cccccccc}
       \toprule
        \multirow{2}{*}{Task} & \multirow{2}{*}{Variant} & \multicolumn{3}{c}{FastDownward} & \multicolumn{3}{c}{ENHSP} \\
        \cmidrule(lr){3-5} \cmidrule(lr){6-8} 
         & & Transl. (s) & Search (ms) & Total (s) & Ground (s) & Search (ms) & Total (s)\\
        \midrule
        \multirow{3}{*}{\texttt{move}} & Easy & $29.55 \pm 0.07$ & $9.05 \pm 3.69$ & $29.80 \pm 0.42$ &  $9.75 \pm 0.02$ & $21.60 \pm 0.37$ & $\mathbf{25.19 \pm 0.48}$ \\
         & Medium & --- & --- &  --- & $49.05 \pm 0.48$ & $\timeout$ & $\timeout$ \\
         & Hard  & --- & --- & --- & --- & --- & --- \\
        \midrule
        \multirow{3}{*}{\makecell{\texttt{pickup}\\\texttt{diamond}}} & Easy & $242.85\pm{0.68}$ & $172.88\pm{1.62}$ & $\mathbf{246.46\pm{0.65}}$ &  $9.28 \pm 0.09$ & $\num{7.15e6} \pm \num{0.32e6}$ & $\num{7.19e6} \pm \num{0.31e6}$ \\
         & Medium & --- & --- & --- & $47.32 \pm 0.29$ & $\timeout$ & $\timeout$ \\
         & Hard  & --- & --- & --- & --- & --- & --- \\
        \midrule
        \multirow{3}{*}{\makecell{\texttt{gather}\\\texttt{wood}}} & Easy & $248.96\pm{0.90}$ & $511.03\pm{3.0}$ & $\mathbf{251.35\pm{0.93}}$ &  $7.42 \pm 0.42$ & $4\,352.49 \pm 76.25$ & $4\,382.79 \pm 77.91$ \\
         & Medium &--- & --- &  --- & $47\,32 \pm 0.36$ & $\timeout$ & $\timeout$ \\
         & Hard  & --- & --- & --- & --- & --- & --- \\
        \midrule
        \multirow{3}{*}{\makecell{\texttt{place}\\\texttt{wood}}} & Easy & $608.85\pm{2.62}$ & $85420\pm{28.4}$ & $\mathbf{973.33\pm{2.56}}$ &  $7.53 \pm 0.07$ & $\timeout$ & $\timeout$ \\
         & Medium & --- & --- &  --- & $49.83 \pm 0.23$ & $\timeout$ & $\timeout$ \\
         & Hard  & --- & --- & --- & --- & --- & --- \\
        \midrule
        \multirow{3}{*}{\makecell{\texttt{pickup and}\\\texttt{place}}} & Easy &   $597.39\pm{1.88}$ & $525\,235\pm{392}$ &  $\mathbf{1\,134.46\pm{4.97}}$ & $7.36 \pm 0.27$ & $\timeout$ & $\timeout$ \\
         & Medium & --- & --- &  --- & $50.23 \pm 0.68$ & $\timeout$ & $\timeout$ \\
         & Hard  & --- & --- & --- & --- & --- & --- \\
        \midrule
        \multirow{3}{*}{\makecell{\texttt{gather multi}\\\texttt{wood}}} & Easy & --- & --- & --- &  $7.62 \pm 0.09$ & $\timeout$ & $\timeout$ \\
         & Medium & --- & --- &  --- & $50.91 \pm 0.72$ & $\timeout$ & $\timeout$ \\
         & Hard  & --- & --- & --- & --- & --- & --- \\
        \midrule
        \multirow{3}{*}{\makecell{\texttt{climb}}} & Easy & --- & --- & --- &  $8.70 \pm 0.07$ & $\timeout$ & $\timeout$ \\
         & Medium & --- & --- &  --- & $48.71 \pm 0.14$ & $\timeout$ & $\timeout$ \\
         & Hard  & --- & --- & --- & --- & --- & --- \\
        \midrule
        \multirow{3}{*}{\makecell{\texttt{cut}\\\texttt{tree}}} & Easy & --- & --- & --- &  $60.35 \pm 1.38$ & $\timeout$ & $\timeout$ \\
         & Medium & --- & --- &  --- & --- & --- & --- \\
         & Hard  & --- & --- & --- & --- & --- & --- \\
        \midrule
        \multirow{3}{*}{\makecell{\texttt{build}\\\texttt{bridge}}} & Easy & --- & --- & --- &  $13.38 \pm 0.13$ & $\timeout$ & $\timeout$ \\
         & Medium & --- & --- &  --- & $57.10 \pm 2.96$ & $\timeout$ & $\timeout$ \\
         & Hard  & --- & --- & --- & --- & --- & --- \\
        \midrule
        \multirow{3}{*}{\makecell{\texttt{build}\\\texttt{cross}}} & Easy & --- & --- & --- &  $8.31 \pm 0.72$ & $\timeout$ & $\timeout$ \\
         & Medium & --- & --- &  --- & $48.34 \pm 1.68$ & $\timeout$ & $\timeout$ \\
         & Hard  & --- & --- & --- & --- & --- & --- \\
        \midrule
        \multirow{3}{*}{\makecell{\texttt{build}\\\texttt{wall}}} & Easy & --- & --- & --- &  $8.48 \pm 0.10$ & $\timeout$ & $\timeout$ \\
         & Medium & --- & --- &  --- & $50.43 \pm 1.54$ & $\timeout$ & $\timeout$ \\
         & Hard  & --- & --- & --- & --- & --- & --- \\
        \midrule
        \pagebreak
        \multirow{3}{*}{\makecell{\texttt{build}\\\texttt{well}}} & Easy & --- & --- & --- &  $10.74 \pm 0.08$ & $\timeout$ & $\timeout$ \\
         & Medium & --- & --- &  --- & $54.40 \pm 1.21$ & $\timeout$ & $\timeout$ \\
         & Hard  & --- & --- & --- & --- & --- & --- \\
        \midrule
        \multirow{3}{*}{\makecell{\texttt{build}\\\texttt{shape}}} & Easy & --- & --- & --- &  $8.06 \pm 0.83$ & $\timeout$ & $\timeout$ \\
         & Medium & --- & --- &  --- & $51.70 \pm 0.72$ & $\timeout$ & $\timeout$ \\
         & Hard  & --- & --- & --- & --- & --- & --- \\
        \midrule
        \multirow{3}{*}{\makecell{\texttt{collect and}\\\texttt{build shape}}} & Easy & --- & --- & --- &  $8.19 \pm 0.13$ & $\timeout$ & $\timeout$ \\
         & Medium & --- & --- &  --- & $51.10 \pm 2.31$ & $\timeout$ & $\timeout$ \\
         & Hard  & --- & --- & --- & --- & --- & --- \\
        \midrule
        \multirow{3}{*}{\makecell{\texttt{build}\\\texttt{cabin}}} & Easy & --- & --- & --- &  ---& ---& ---\\
         & Medium & --- & --- &  --- & ---& ---& ---\\
         & Hard  & --- & --- & --- & --- & --- & --- \\
        \bottomrule
    \end{tabular}}
    }
\caption{The running times for Fast Downward and ENHSP-20 when run on the MinePlanner task suite. For results marked as ---, no result could be obtained because the planner failed to translate (in the case of Fast Downward) or ground (for ENHSP-20). Entries marked $\timeout$ indicate that the planner timed out. Means and standard deviations over 5 runs are reported.}
\label{tab:full-results}
\end{table}

\end{document}


\maketitle

\onecolumn

\section{Examples of Operators}

The following are examples of operators for moving north for two styles of planners. The precondition is this operator describes how there must be no block at the location the agent wishes to move to (since this would obstruct the agent) and that no item exists either, since the effect would then include picking up the item. The effect in this case is simply a change in the agent's $z$ position. 

\begin{listing*}[h!]%
\caption{Example of a movement action using fluents}%
\label{lst:pddl_num_movement}%
\begin{lstlisting}[language=Lisp]
(:action move-north
 :parameters (?ag - agent)
 :precondition (and (agent-alive ?ag) 
  (exists (?b - block) (and  (block-present ?b) (= (x ?b) (x ?ag)) 
  (= (y ?b) (+ (y ?ag) -1)) (= (z ?b)  (+ (z ?ag) -1)))) (and 
  (not (exists (?b - block) (and  (block-present ?b) (= (x ?b) (x ?ag)) 
  (or (= (y ?b) (+ (y ?ag) 1)) (= (y ?b) (y ?ag))) (= (z ?b) (+ (z ?ag) -1))))) 
  (not (exists (?i - item) (and (item-present ?i) (= (x ?i) (x ?ag)) (= (y ?i) (y ?ag)) (= (z ?i) (+ (z ?ag) -1)))))))
 :effect (and (decrease (z ?ag) 1))
)\end{lstlisting}
\end{listing*}

\begin{listing*}[h!]%
\caption{Example of a movement action using predicates}%
\label{lst:pddl_pred_movement}%
\begin{lstlisting}[language=Lisp]
(:action move-north
 :parameters (?ag - agent ?x - position ?y_up - position ?y_down - position ?y_2_down - 
  position ?z_start - position ?z_end - position)
 :precondition (and (agent-alive ?ag) (at-x ?ag ?x) (at-y ?ag ?y_down)
    (at-z ?ag ?z_start) (are-seq ?z_end ?z_start) (are-seq ?y_down ?y_up) (are-seq ?y_2_down ?y_down) (exists (?b - block) (and (block-present ?b) (at-x ?b ?x) (at-y ?b ?y_2_down) (at-z ?b ?z_end))) (not (exists (?b - block) (and (block-present ?b) (at-x ?b ?x) (or (at-y ?b ?y_up)  (at-y ?b ?y_down)) (at-z ?b ?z_end)))) (not (exists (?i - item) (and (item-present ?i) (at-x ?i ?x) (at-y ?i ?y_down) (at-z ?i ?z_end)))))
 :effect (and (not (at-z ?ag ?z_start)) (at-z ?ag ?z_end))
)
\end{lstlisting}
\end{listing*}

The listings below illustrate an example of the agent interacting with a block. Here, the agent breaks a grass block in the north direction. The precondition checks that there is a grass block north of the agent (and no other items that could be picked up mistakenly), and the effect is that the block is no longer in the world, and the count for the agent's inventory of grass blocks increases by 1.

\begin{listing*}[h!]%
\caption{Example of an interaction action using fluents}%
\label{lst:pddl_num_movement}%
\begin{lstlisting}[language=Lisp]
(:action break-grass_block-north
 :parameters (?ag - agent ?b - grass_block-block)
 :precondition (and 
    (= (x ?b) (x ?ag)) (= (y ?b) (y ?ag)) (= (z ?b) (+ (z ?ag) -1)) (block-present ?b) (not (exists (?i - item) (and (item-present ?i) (= (x ?b) (x ?ag)) (= (y ?i) (+ (y ?ag) 1)) (= (z ?b) (+ (z ?ag) -1))))))
 :effect (and (not (block-present ?b)) (increase (agent-num-grass_block ?ag) 1))
)
)\end{lstlisting}
\end{listing*}

\begin{listing*}[h!]%
\caption{Example of an interaction action using predicates}%
\label{lst:pddl_pred_movement}%
\begin{lstlisting}[language=Lisp]
(:action break-grass_block-north
 :parameters (?ag - agent ?b - grass_block-block ?x - position ?y - position ?y_up - position ?z - position ?z_front - position ?n_start - count ?n_end - count)
 :precondition (and 
    (agent-alive ?ag) (at-x ?ag ?x) (at-y ?ag ?y) (at-z ?ag ?z) (at-x ?b ?x) (at-y ?b ?y)
     (at-z ?b ?z_front) (are-seq ?z_front ?z) (are-seq ?y ?y_up) (block-present ?b) (not (exists (?i - item) (and (item-present ?i)(at-x ?i ?x) (at-y ?i ?y_up) (at-z ?i ?z_front)))) 
     (are-seq ?n_start ?n_end) (agent-has-n-grass_block ?ag ?n_start)
    )
 :effect (and (not (block-present ?b)) (not (at-x ?b ?x)) (not (at-y ?b ?y)) (not (at-z ?b ?z_front)) (not (agent-has-n-grass_block ?ag ?n_start)) (agent-has-n-grass_block ?ag ?n_end)
    )
)
\end{lstlisting}
\end{listing*}

\pagebreak


\section{Results}

We benchmark Fast Downward, a propositional planner, and ENHSP-20, a numerical planner, on the tasks provided with \mineplanner.
All experiments were conducted using a cluster of AMD Ryzen 9 7900X3D CPUs, using 128 virtual cores and 250GB of RAM per trial.
We record the time taken for planning, including the amount of time spent preprocessing the PDDL file by each planner. For Fast Downward, this is refers to the time taken to translate PDDL to SAS, while for ENHSP, this measures grounding.  We set a timeout limit of 2 hours for each planing task, since an autonomous agent must ultimately be capable of planning in large environments within a reasonable timeframe.
Results are reported in Table 1, with the means and standard deviations over 5 runs reported.

 \begin{adjustbox}{angle=90,captionbelow={The running times for Fast Downward and ENHSP-20 when run on the MinePlanner task suite. For results marked as ---, no result could be obtained because the planner failed to translate (in the case of Fast Downward) or ground (for ENHSP-20). Entries marked $\timeout$ indicate that the planner timed out.}, float=table}
\renewcommand{\arraystretch}{0.4}
\centering
    \begin{tabular}{cccccccc}
       \toprule
        \multirow{2}{*}{Task} & \multirow{2}{*}{Variant} & \multicolumn{3}{c}{FastDownward} & \multicolumn{3}{c}{ENHSP} \\
        \cmidrule(lr){3-5} \cmidrule(lr){6-8} 
         & & Transl. (s) & Search (ms) & Total (s) & Ground (s) & Search (ms) & Total (s)\\
        \midrule
        \multirow{3}{*}{\texttt{move}} & Easy & $41.80 \pm 0.15$ & $6.80 \pm 0.55$ & $41.91 \pm 0.17$ &  $11.64 \pm 0.13$ & $20.20 \pm 0.40$ & $20.40 \pm 0.30$ \\
         & Medium & $237.69 \pm 0.86$ & $14.60 \pm 5.88$ &  $237.68 \pm 0.71$ & $73.43 \pm 0.94$ & $\num{168e3} \pm \num{6.9e3}$ & $317.20 \pm 5.92$ \\
         & Hard  & --- & --- & --- & --- & --- & --- \\
        \midrule
        \multirow{3}{*}{\makecell{\texttt{pickup}\\\texttt{diamond}}} & Easy & $341.21 \pm 2.12$ & $497.36 \pm 3.67$ & $341.96 \pm 2.08$ &  $11.83 \pm 0.14$ & $45.40 \pm 1.36$ & $\mathbf{20.52 \pm 0.14}$ \\
         & Medium & --- & --- & --- & $72.18 \pm 0.34$ & $49\,676 \pm 919$ & $\mathbf{196.24 \pm 1.72}$ \\
         & Hard  & --- & --- & --- & --- & --- & --- \\
        \midrule
        \multirow{3}{*}{\makecell{\texttt{gather}\\\texttt{wood}}} & Easy & $343.76 \pm 0.66$ & $481.64 \pm 1.45$ & $344.43 \pm 0.66$ &  $12.26 \pm 0.75$ & $1125 \pm 37.26$ & $\mathbf{25.26 \pm 0.91}$ \\
         & Medium &--- & --- &  --- & $71.82 \pm 0.67$ & $12\,065 \pm 144$ & $\mathbf{157.05 \pm 3.17}$ \\
         & Hard  & --- & --- & --- & --- & --- & --- \\
        \midrule
        \multirow{3}{*}{\makecell{\texttt{place}\\\texttt{wood}}} & Easy & $786.31 \pm 2.03$ & $7528 \pm 28.1$ & $\mathbf{792.21 \pm 1.98}$ &  $12.81 \pm 0.03$ & $\timeout$ & $\timeout$ \\
         & Medium & --- & --- &  --- & $58.49 \pm 0.23$ & $\timeout$ & $\timeout$ \\
         & Hard  & --- & --- & --- & --- & --- & --- \\
        \midrule
        \multirow{3}{*}{\makecell{\texttt{pickup and}\\\texttt{place}}} & Easy &  $341.21 \pm 2.12$ & $993.7 \pm 3.33$ &  $\mathbf{343.94 \pm 1.52}$ & $11.53 \pm 0.11$ & $\num{1.63e6} \pm \num{0.006e6}$ & $1655 \pm 60.4$ \\
         & Medium & --- & --- &  --- & $73.79 \pm 0.85$ & $\timeout$ & $\timeout$ \\
         & Hard  & --- & --- & --- & --- & --- & --- \\
        \midrule
        \multirow{3}{*}{\makecell{\texttt{gather multi}\\\texttt{wood}}} & Easy & --- & --- & --- &  $11.76 \pm 0.075$ & $241\,840 \pm 7150$ & $\mathbf{256.86 \pm 7.78}$ \\
         & Medium & --- & --- &  --- & $75.91 \pm 0.72$ & $\timeout$ & $\timeout$ \\
         & Hard  & --- & --- & --- & --- & --- & --- \\
        \midrule
        \multirow{3}{*}{\makecell{\texttt{climb}}} & Easy & --- & --- & --- &  $15.52 \pm 0.07$ & $\timeout$ & $\timeout$ \\
         & Medium & --- & --- &  --- & $57.51 \pm 0.14$ & $\timeout$ & $\timeout$ \\
         & Hard  & --- & --- & --- & --- & --- & --- \\
        \midrule
        \multirow{3}{*}{\makecell{\texttt{cut}\\\texttt{tree}}} & Easy & --- & --- & --- &  $72.90 \pm 1.21$ & $\timeout$ & $\timeout$ \\
         & Medium & --- & --- &  --- & --- & --- & --- \\
         & Hard  & --- & --- & --- & --- & --- & --- \\
        \midrule
        \multirow{3}{*}{\makecell{\texttt{build}\\\texttt{bridge}}} & Easy & --- & --- & --- &  $16.26 \pm 0.08$ & $\timeout$ & $\timeout$ \\
         & Medium & --- & --- &  --- & $84.29 \pm 3.01$ & $\timeout$ & $\timeout$ \\
         & Hard  & --- & --- & --- & --- & --- & --- \\
        \midrule
        \multirow{3}{*}{\makecell{\texttt{build}\\\texttt{cross}}} & Easy & --- & --- & --- &  $15.04 \pm 0.82$ & $\timeout$ & $\timeout$ \\
         & Medium & --- & --- &  --- & $81.24 \pm 2.13$ & $\timeout$ & $\timeout$ \\
         & Hard  & --- & --- & --- & --- & --- & --- \\
        \midrule
        \multirow{3}{*}{\makecell{\texttt{build}\\\texttt{wall}}} & Easy & --- & --- & --- &  $13.20 \pm 0.10$ & $\timeout$ & $\timeout$ \\
         & Medium & --- & --- &  --- & $79.54 \pm 1.54$ & $\timeout$ & $\timeout$ \\
         & Hard  & --- & --- & --- & --- & --- & --- \\
        \midrule
        \pagebreak
        \multirow{3}{*}{\makecell{\texttt{build}\\\texttt{well}}} & Easy & --- & --- & --- &  $13.04 \pm 0.08$ & $\timeout$ & $\timeout$ \\
         & Medium & --- & --- &  --- & $81.24 \pm 2.13$ & $\timeout$ & $\timeout$ \\
         & Hard  & --- & --- & --- & --- & --- & --- \\
        \midrule
        \multirow{3}{*}{\makecell{\texttt{build}\\\texttt{shape}}} & Easy & --- & --- & --- &  $10.44 \pm 0.83$ & $\timeout$ & $\timeout$ \\
         & Medium & --- & --- &  --- & $61.32 \pm 0.25$ & $\timeout$ & $\timeout$ \\
         & Hard  & --- & --- & --- & --- & --- & --- \\
        \midrule
        \multirow{3}{*}{\makecell{\texttt{collect and}\\\texttt{build shape}}} & Easy & --- & --- & --- &  $12.39 \pm 0.06$ & $\timeout$ & $\timeout$ \\
         & Medium & --- & --- &  --- & $68.17 \pm 1.98$ & $\timeout$ & $\timeout$ \\
         & Hard  & --- & --- & --- & --- & --- & --- \\
        \midrule
        \multirow{3}{*}{\makecell{\texttt{build}\\\texttt{cabin}}} & Easy & --- & --- & --- &  ---& ---& ---\\
         & Medium & --- & --- &  --- & ---& ---& ---\\
         & Hard  & --- & --- & --- & --- & --- & --- \\
        \bottomrule
    \end{tabular}
        \label{tab:results}
\end{adjustbox}

